\pdfoutput=1

\documentclass[11pt]{article}

\usepackage[table,xcdraw]{xcolor}
\usepackage{acl}

\usepackage{times}
\usepackage{latexsym}

\usepackage[T1]{fontenc}

\usepackage[utf8]{inputenc}

\usepackage{microtype}

\usepackage{inconsolata}

\usepackage{CJKutf8}

\usepackage[utf8]{inputenc}

\usepackage{graphicx}
\usepackage{amsmath}
\usepackage{amsfonts}
\usepackage{amssymb}
\usepackage{url}

\title{A Measure for Transparent Comparison of Linguistic Diversity in Multilingual NLP Data Sets}

 \author{Tanja Samardžić \\ Language and Space Lab \\ University of Zurich \\ \texttt{tanja.samardzic@uzh.ch} \And 
 Ximena Gutierrez \\  CEIICH \\  Universidad Nacional \\ Autónoma de México \And
 Christian Bentz \\   Dept. of General Linguistics  \\  Eberhard-Karls-Universität \\ Tübingen \AND
 Steven Moran \\ Laboratory of Language Evolution \\ University of Neuchâtel \And
Olga Pelloni       \\ Telepathy Labs GmbH \\ Zurich, Switzerland}

\begin{document}
\maketitle
\begin{abstract}


Typologically diverse benchmarks are increasingly created to track the progress achieved in multilingual NLP. Linguistic diversity of these data sets is typically measured as the number of languages or language families included in the sample, but such measures do not consider structural properties of the included languages. In this paper, we propose assessing linguistic diversity of a data set against a reference language sample as a means of maximising linguistic diversity in the long run. We represent languages as sets of features and apply a version of the Jaccard index ($J_{mm}$) suitable for comparing sets of measures. In addition to the features extracted from typological data bases, we propose an automatic text-based measure, which can be used as a means of overcoming the well-known problem of data sparsity in manually collected features. Our diversity score is interpretable in terms of linguistic features and can identify the types of languages that are not represented in a data set. Using our method, we analyse a range of popular multilingual data sets (UD, Bible100, mBERT, XTREME, XGLUE, XNLI, XCOPA, TyDiQA, XQuAD). In addition to ranking these data sets, we find, for example, that (poly)synthetic languages are missing in almost all of them.

\end{abstract}

\section{Introduction}

\begin{figure*}
\begin{center}
        \includegraphics[width=1\textwidth]{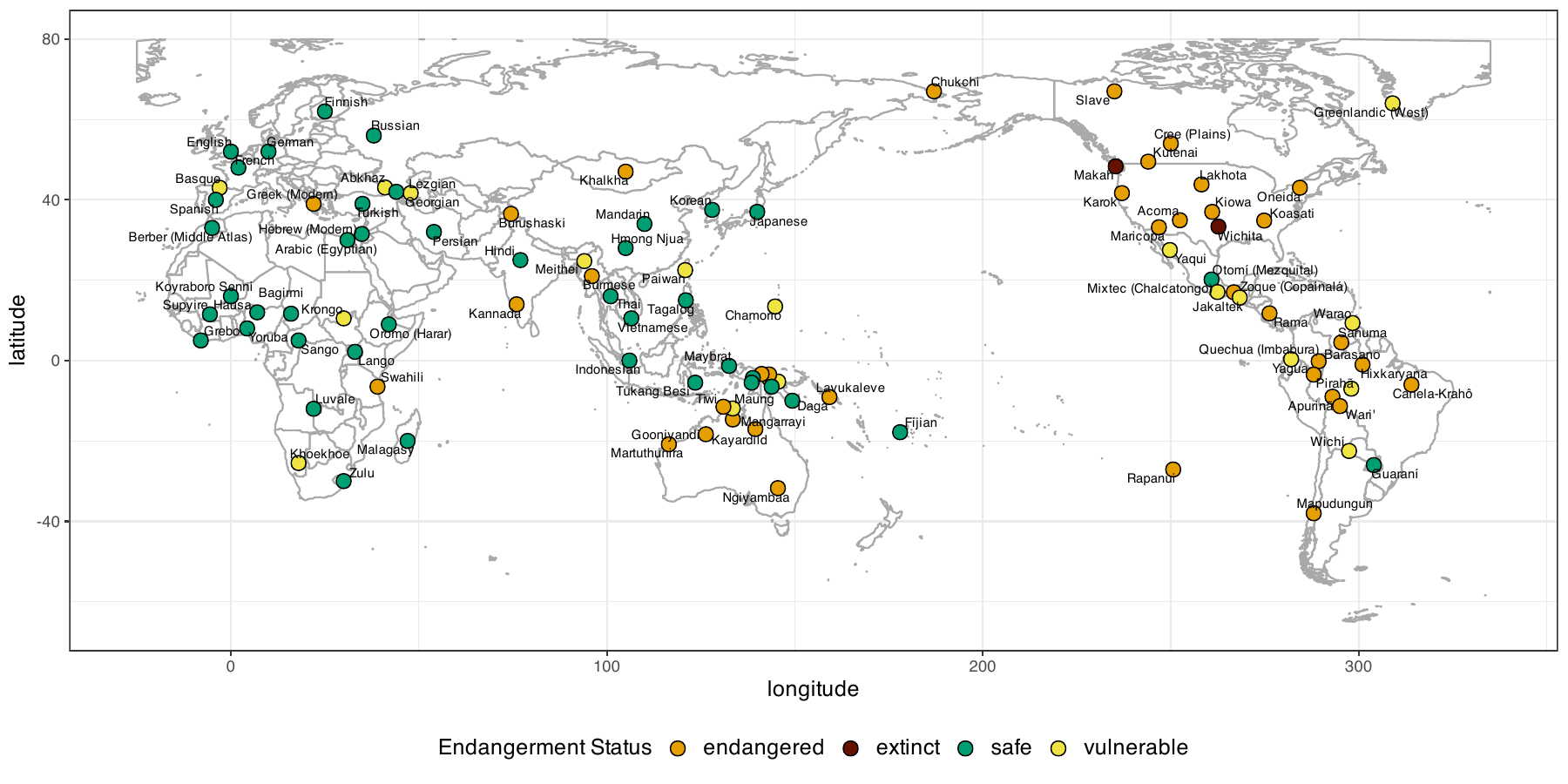} 
\caption{Languages in the WALS 100L sample with their endangerment status.}
\label{100LC-sample}
\end{center}
\end{figure*}



Data sets for training and testing NLP models are increasingly multilingual and aimed at broad linguistic coverage. These data sets are often claimed to represent a typologically diverse sample, including low-resource and endangered languages.

Linguistic diversity is typically described as the number of languages included in the data set, yet less often as the number of language families to which these languages belong. Both counts indicate a level of linguistic diversity: the more languages and families, the more diversity. But how do we know that included languages are indeed different? How can we define a desired or optimal diversity to set as a goal when composing multilingual data sets? 
These questions need to be addressed if our goal is to know how NLP technology generalises across diverse languages, without testing it on each single language (even if we had the necessary data for all languages).

The aim of this paper is to initiate and facilitate comparisons between multilingual NLP data sets with respect to a linguistic diversity reference. For this, we propose a measure of linguistic diversity and a method of comparison that identifies what kinds of linguistic features are missing. As an initial reference, we rely on a predefined sample of languages --- the 100-language-sample (100L) selected by the Word Atlas of Language Structures (WALS; \citet{Comrie2013}) to represent geographic and phylogenetic diversity.  As a comparison method, we formulate a version of the Jaccard index suitable for comparing measures. This measure allows us to quantify the distance between the observed and the reference diversity in terms of linguistic features, showing not only how diverse language samples are but also what kinds of linguistic phenomena are not represented in a given sample. 
To facilitate automatic extraction of linguistic features needed for assessing linguistic diversity, we complement the information from linguistic data bases with relevant text statistics. 

Our proposals are intended to help researchers make informed choices when designing a multilingual data set. Representing a wider spectrum of linguistic diversity is not only a way to improve the cross-linguistic generalisation of NLP technology, but also a way to deal with biases against low-resource languages, which are harder to represent and thus more likely to be left behind \citep{joshi-etal-2020-state}.  

\section{Background and Related Work}
\label{sec:related-work}

Evaluating the linguistic diversity of data sets relies on comparable descriptions of languages. For instance, the (approximate) number of speakers is an attribute whose value can be found and compared for all registered languages. This attribute, however, does not describe the structure of languages. An example of a structural attribute would be the presence or the absence of adjectives in a language. To establish the value of this attribute for any language, we need a universal definition of what an adjective is. It turns out that such universal definitions are hard to formulate in a principled way \citep{haspelmath_2007_comparison}, which makes it hard to define objective measures of how similar or dissimilar any two languages are. 

The most widely accepted method for comparing languages relies on genealogical classification: given a phylogenetic tree, we consider languages located in the same region of the tree to be similar. This method currently prevails in NLP (cf.\ the work discussed in Section \ref{sec:results}). Typically, we regard languages that belong to the same \textit{family} to be similar. To know which language belongs to which family, we turn to popular authorities such as WALS \citep{wals} or Glottolog \citep{glottolog}.
However, language families can be too broad for a meaningful comparison as they include typologically very different languages. For instance, English and Armenian belong to the same family, Indo-European, but are vastly different in terms of their phoneme inventories, morphology, and word order. 

Another possibility to compare languages, starting to be used in NLP only recently, is to rely on grammatical features available in the WALS data base.  This is a comprehensive source of information about linguistic structures but still rather sparsely populated;  feature values are often known for only a few languages.\footnote{An alternative typological data base is AUTOTYP \citep{bickel2017}, with a different design but similar coverage.} 
Together with other typological data bases, WALS is included in URIEL, an aggregated and standardised source of language features for various NLP uses.  \citet{ponti-etal-2020-xcopa} propose a diversity score using the features from URIEL \citep{littell-etal-2017-uriel}. The score is called \textit{typology index} and it is calculated as the entropy of feature values (averaged per data set).\footnote{They propose two more scores, \textit{family} and \textit{geography}, which do not make use of grammatical features.} In other NLP work, grammatical features (usually termed \textit{typological}) are used for other purposes, such as predicting the features \citep{ponti-etal-2019-modeling} rather than using them for language sampling in creating multilingual data sets. \citet{moran-2016-acqdiv} use WALS and AUTOTYP features \cite{Stolletal2013Capturing} to compose a sample of 10 maximally diverse languages for a corpus-based study of language acquisition. 

Finally, languages can be described using features derived from various text statistics, but such features are not commonly used for language sampling. Type-token ratio (TTR) or unigram entropy of a text have been shown to correlate with grammar-based morphological complexity measures \citep{kettunen2014, bentz2016}. 
Many other methods have been proposed for assessing linguistic complexity using text statistics (see, for instance, \citet{berdicevskis2018}). 
All of these measures can, in principle, be used for describing and comparing languages although such comparisons might seem counter-intuitive and hard to interpret in terms of genealogical classification.  On the other hand, these features might  complement usual descriptions of languages while being more directly relevant to text processing and NLP.

Transfer learning created a new need for nuanced languages comparison for NLP. While models can now be transferred across languages with zero-shot or few-shot learning \citep{pires-etal-2019-multilingual}, the success of the transfer might depend on the differences between languages. \citet{lin-etal-2019-choosing} propose a range of measures that can be used in order to choose the best transfer language, which they divide into data-dependent (data size, token overlap, TTR) and data independent (various distance measures extracted from the URIEL data base).  \citet{lauscher-etal-2020-zero} study how well different similarity scores predict the success of the transfer and they find that language family is, in fact, the one that is least helpful in all the tasks considered (with mBERT and XLM-R). Various criteria for assessing language similarity remain an open research area in NLP \citep{turc2021revisiting,pelloni-etal-2022-subword,samardzic-etal-2022-language,de-vries-etal-2022-make}. Our proposal for assessing linguist diversity is relevant to these efforts too, as its key component is language comparison at the level of features extracted from both typological data bases and text samples.  

More generally, our work is intended to contribute to several wide-scope initiatives for improving the quality of data management in multilingual NLP \citep{bender-friedman-2018-data,kreutzer2021quality,lhoest-etal-2021-datasets} by focusing specifically on diversity assessments and data-independent scores for language comparison.


\section{Comparing Data Sets with Jaccard Similarity}

\begin{figure}
    \centering
    \resizebox{6cm}{!}{  
    \includegraphics[width=.48\textwidth]{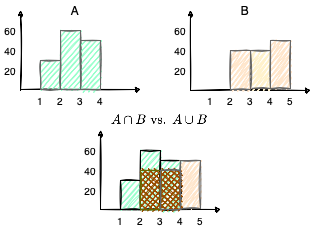}
    }   
    \caption{A toy example of comparing sets of measures with the minmax version of the Jaccard index.}
    \label{fig:ex_Jaccard}
\end{figure}

Our goal is to estimate the linguistic diversity of a data set with respect to some reference. Our score is thus a comparison between two data sets. More precisely, we compare scaled distributions of the values of a numerical attribute as shown in Figure~\ref{fig:ex_Jaccard}. The upper part of the figure shows (constructed) examples of two data sets (A and B), which we compare assuming that A is the data set whose diversity we want to assess and B is the reference. The values of the numerical attribute (one measurement per language) are on the x-axis and the numbers of languages are on the y-axis. Each bar in the figures represents the number of languages in the given data set with the numerical value in the given range (bin). For instance, the first bar in the upper left plot shows that the first sample (A) has 30 languages, with the values of their numerical attributes falling between 1 and 2. The other sample (B) has no languages in this bin. 

The width of the bins is arbitrary, but it does impact the score. Narrower bins capture more differences between two distributions than wider bins. By setting the width of the bins, we thus control the resolution at which we want to compare two data sets. In our example, the width is the distance between integers, but one can define other values (as long as the bins are of the same width).    

Since the data sets that we compare contain different numbers of languages, the values on the y-axis (counts of languages) are normalised in order to neutralise the effect of the size of the samples and focus rather on the diversity.
We multiply all counts in the smaller set with the scalar $c$: 

\begin{equation}
    c = \frac{max(|A|,|B|)}{min(|A|,|B|)}
\end{equation}

In this way, we increase the counts in the smaller set proportionally to obtain the same number of data points in both distributions and comparable numbers in each bin.\footnote{Another way to normalise the counts would be to divide them by the size of the set, but we chose the first option in order to preserve the notion of \textit{number of languages}, which is helpful for the subsequent explanations.} 

Once we have represented our two sets in this way, we compare them using a generalised version of Jaccard similarity. This score shows how much the two distributions overlap. 
The original Jaccard index \citep{JaccardTHEDO} compares two sets, but its generalised versions are suitable for comparing sets of measurements. Thus, we use the \textit{minmax} version of the score ($J_{mm}$), initially proposed by \citet{Tanimoto} for comparing vectors of binary values and then generalised to weight vectors by \citet{Grefenstette-1994}. In our version, we compare two data sets as two vectors of weights: each bin 
is one dimension in the vectors and the number of languages in that bin is its weight. 

Intuitively, the score is the ratio between the intersection and the union of the two distributions (shown in the bottom part of Figure \ref{fig:ex_Jaccard}). Formally, we first map all the languages in all data sets to real numbers $m: \Bbb{L} \mapsto \Bbb{R}$, so that $\{Y = m(x): x \in X\} = \{(x_i,y_i)\}$, where $x$ is a language in a data set, $y$ is its corresponding measurement ($y \in \Bbb{R}$) and the range of the index $i$ $1\dots |X|$ is the set of languages included in a data set. We then group the measurements into bins by applying a given threshold: $\{Z = t(y): y \in Y\} = \{(y_i,z_j)\}$, where $z$ is the bin to which the measurement is assigned, the range of $i$ is  $1\dots |X|$  and the range of $j$ is $1\dots |Z|$.

With this formalisation, we define the Jaccard minmax similarity of two data sets, $J_{mm}(A,B)$, as a similarity between two vectors of weights:

\begin{equation}
    J_{mm}(\mathbf{a},\mathbf{b}) = \frac{\sum_{j=1}^{|Z|}min(a_j,b_j)}{\sum_{j=1}^{|Z|}max(a_j,b_j)} 
\end{equation}

The sum in the numerator represents the intersection and the sum in the denominator the union of the two sets of measurements. The weights $a$ and $b$ represent the number of measurements in the bin $j$. The values of $J_{mm}$ fall in the range $[0,1]$, with higher values indicating more similarity between A and B, and, indirectly, better coverage of linguistic diversity in A. 






What is especially interesting about using $J_{mm}$ as a diversity score is its transparency in terms of individual measurements: we can visualise and interpret where exactly a data set departs from the reference. 

\section{Language Features}
\label{sec:features}

We now turn to the question of how to define and take measures (the values on the x-axis in Figure \ref{fig:ex_Jaccard}) that can be used for calculating Jaccard minmax similarity between sets of languages. We use two kinds of descriptions.


\subsection{Grammar Features}
Typological data bases are currently the principal source of information about the properties of languages, but NLP researchers are faced with many obstacles when using this information. The popular software package \texttt{lang2vec}  associated with the URIEL data base \citep{littell-etal-2017-uriel} alleviates some of the obstacles. First, the package solves the problem of incompatible feature values across different sources by mapping the data from several  original data bases to binary features. Second, the problem of sparsity of feature values is solved by imputing the missing values: instead of a missing feature value in a language, the package returns the observed value for the same feature in the closest language. In this way, features become available for all queried languages, which is necessary for estimating language diversity, but a large proportion (roughly 40\%) of the returned features are imputed. 

While \texttt{lang2vec} facilitates retrieving typological features, its use for describing languages is limited due to remaining obstacles that are hard to solve. First, it does not contain any morphological features, which are especially relevant to NLP due to the known difficulties with that morphologically rich languages \citep{tsarfaty-etal-2013-parsing}. The second unsolved problem is the fact that  typological features are hard to add for languages for which they are not already available. Adding new features requires human expertise in many languages. 


\subsection{Text Features}
As a complement to commonly used features from \texttt{lang2vec}, we make use of linguistically relevant text statistics. In this study, we focus on the \textit{mean word length} as an approximation of aggregated morphological features, but other text-based features might be envisioned in future work. Our choice to start with word length relies on the observation that longer words can be expected in languages with rich morphology (large morphological paradigms, productive derivation), while shorter words are found in languages with less morphology.\footnote{We give a more specific definition of the notion of a word as part of the methods in Section \ref{sec:methods}.} As an empirical confirmation of the relationship between the word length and morphology, we perform a correlation test between the mean word length and morphological complexity calculated over morphological features (see Section \ref{sec:methods} for the methods and Section \ref{sec:discussion} for the discussion).

Text features are especially interesting in the context of NLP because they can be calculated automatically and applied to any language in which there are any texts to process. An important advantage of word length over other text statistics in this regard is that it manifests itself in very small samples of text and remains stable across different sizes. A sample of contiguous text of only 500 tokens gives us already a very good estimation of the overall mean word length. This can be seen in Figure \ref{fig:ex_mwl100LC} in the Appendix \ref{sec:appendixcorr}, which shows the values of the mean word length on random samples of the length 500, 2000 and 10000 tokens in 87 languages. A correlation score (also in the Appendix \ref{sec:appendixcorr}) shows that languages are almost identically ranked with all the sample sizes.

\subsection{Maximising Linguistic Diversity}

The editors of the WALS data base have selected two samples of languages (100 and 200 sample) as a means of guidance in the collective effort to create linguistic descriptions on a wide scale. These samples maximise genealogical (language family) and areal (geographic) diversity. Completing their descriptions is expected to minimise a potential bias regarding the relative frequency of different types of linguistic features  included in the data base \cite{Comrie2013}.  Figure \ref{100LC-sample} shows the locations of the languages in the 100 sample and their endangerment status according to UNESCO. 

Recently, text samples have been collected for most of the 100 languages in the TeDDi data set \citep{moran-etal-2022-teddi}.\footnote{\url{https://github.com/MorphDiv/TeDDi_sample/tree/master}}
These text data are sampled from online resources, e.g., Project Gutenberg,\footnote{\url{https://www.gutenberg.org/}} Open Subtitles \cite{LisonTiedemann2016}, The Parallel Bible Corpus \cite{Mayer2014}, the Universal Declaration of Human Rights,\footnote{\url{http://unicode.org/udhr/}}, but also from grammars and other language documentation sources. For languages not present in online resources, the texts were manually transcribed. 

We take these two resources as the current reference that maximises linguistic diversity in terms of grammar features (WALS) and text features (TeDDi). We compare NLP data sets with these references, but our method can be applied to compare any given pair of data sets including potentially better references in the future.

\section{Data and Methods}
\label{sec:methods}

We calculate the Jaccard minmax diversity score (J$_{mm}$) for a number of popular multilingual data sets in comparison to the TeDDi sample.\footnote{The code for reproducing the calculations can be found at \url{https://github.com/MorphDiv/jmm_diversity/}.}  Without attempting to provide an exhaustive evaluation, we review data sets that satisfy the following criteria: multilingual (containing ten or more languages), relatively widely used and recently released or updated. The list is given in Table \ref{tab:benchmarks} and discussed in more detail in Section \ref{sec:results}.  For reference, we compare our J$_{mm}$ score to the typological index (TI) previously proposed as a linguistic diversity measure by \citet{ponti-etal-2020-xcopa} (see Section \ref{sec:related-work}).

Descriptions of the data sets often do not include all the information that was needed for our comparison. In particular, the number of language families is often not stated. To add this information, we extracted language names from the data files, converted these names into ISO 639-3 codes manually, and then retrieved the corresponding families from the Glottolog data base (top level family). 
 Note that the conversion to ISO 639-3 codes led to some changes in the number of languages, compared to those cited in the data descriptions. For instance, the mBERT training data has only 97 distinct languages, not 104 as mentioned in the original description.



\subsection{Methods for Text Features}

We define words to be sequences of Unicode characters, delimited by spaces or other language-specific word delimiters, as defined by common multilingual tokenisers.
We tokenise all the collected samples into word-level tokens using the Python library Polyglot \citep{al2015polyglot}.\footnote{\url{https://polyglot.readthedocs.io}} 
If a resulting token does not contain any alphanumeric characters, we discard it as punctuation. All the remaining tokens are further segmented into characters using the Python library \texttt{segments} \citep{Moran2018}.\footnote{\url{https://github.com/cldf/segments}} 
We split words into sequences of characters and take their length as word length.\footnote{We use the units defined by the Unicode Standard as ``user-perceived characters'' (NFC).}  We apply this same definition to all scripts, but we discuss below potential adjustments in the case of (partially) logographic scripts. 

Since the mean word length can be calculated on small samples, we take a single random sample for each language in a data set that we consider. To do this, we select a random position in the data set and extract contiguous text of the length up to 10K tokens starting from the random position. In case a data set does not contain such long texts (or sequences of paragraphs), we take smaller samples. The smallest samples are  200-300 tokens long. 

The output of these text processing steps is a set of real numbers, each number representing a language in a data set. To turn these numbers into discrete features, we group them into bins of equal size. We set the bin width to 1.\footnote{In addition to this, we also tried smaller bin sizes. We do not report the latter results, but the main trends did not change.}

\smallskip
\noindent
\textbf{Mean word length vs. WALS features}
 Following \citet{bentz2016}, we calculate a complexity score ($C_{WALS}$) for each language using the set of 26 features that are relevant to describing morphology. This score is obtained by: 1) transforming the range of values each feature can take so that bigger values reflect the increasing use of morphology; 2) normalizing and averaging the resulting feature values per language. The list of features and transformations is given in Table \ref{tab:WALS} in Appendix \ref{sec:appendixmorph}.   $C_{WALS}$ ranges from 0 to 1, where values closer to one indicate that the language encodes more morphosyntactic distinctions, making its morphology richer. All the values of the mean word length and morphological complexity for 29 diverse languages (the subset of TeDDi languages for which the 26 WALS features are known) are shown in Table \ref{tab:mwlccorr} in Appendix \ref{sec:appendixmorph}. We observe a strong correlation ($\rho = 0.69$), which means that the variables quantify very similar phenomena and that the mean word length is a reasonable approximation of morphological types of languages. We return to this point in Section \ref{sec:discussion}.  

\smallskip
\noindent
\textbf{Adjustments for logographic scripts}
Words in languages with logographic scripts tend to be shorter due to the fact that a single  symbol corresponds to several alphabetic symbols \citep{sproat-logography-2021}. For instance, in Mandarin Chinese, types such as \begin{CJK*}{UTF8}{bsmi}的\end{CJK*} \textit{de} (possessive particle), \begin{CJK*}{UTF8}{bsmi}了\end{CJK*} \textit{le} (aspect particle), \begin{CJK*}{UTF8}{bsmi}是\end{CJK*} \textit{shì} (copular verb `is'),  \begin{CJK*}{UTF8}{bsmi}我們\end{CJK*} \textit{w\v{o}-men} (pronoun `us') are assigned lengths (1, 1, 1, 2) respectively when measured in UTF-8 characters in the original script. When transliterated into Pinyin, the corresponding lengths are (2, 2, 3, 5). Hence, compared to Pinyin, the lengths are somewhat underestimated. It might seem more appropriate to convert the logographic scripts into their romanised counterparts to achieve cross-linguistic comparability. We opt for leaving such scripts without conversion, because we consider this phenomenon part of the diversity that we want to capture. Additional motivation for our choice is the fact that NLP systems have to deal with text as it is regardless of the mapping between written characters and sounds. 

Three languages in our data samples, Chinese, Japanese and Korean, are affected by this issue to a varied degree. 
In these cases, we scale the observed word length proportionally to the difference between the Chinese original script and Pinyin so that the scaled length is comparable to alphabetic scripts. Table \ref{tab:benchmarks_logo} in Appendix \ref{sec:logographic_adjustments} shows revised diversity scores after the adjustments.

\subsection{Linguistic Diversity Scores}


\begin{table*}
\centering
\begin{tabular}{p{6cm}rr|rrrr}
Name and main references                                                                                                                            & N(L) & N(F) & TI$_{syn}$                      & $J_{mm\_syn}$                 & TI$_{morph}$                    & $J_{mm\_morph}$ \\ \hline
Universal Dependencies (UD)                    & 106* & 20*  & \cellcolor[HTML]{F0B06B}0.567   & \cellcolor[HTML]{F0B06B}0.736 & \cellcolor[HTML]{EC8532} 0.349                           & \cellcolor[HTML]{EA771F} 0.650           \\
Bible 100                                                 & 103* & 30*  & \cellcolor[HTML]{EA771F}0.649   & \cellcolor[HTML]{EA771F}0.811 & \cellcolor[HTML]{F3CD91} 0.311                           &  \cellcolor[HTML]{F2BE7E} 0.534           \\
mBERT                                                & 97*  & 15*  & \cellcolor[HTML]{F2BE7E}0.559   & \cellcolor[HTML]{F3CD91}0.710 & \cellcolor[HTML]{F2BE7E} 0.323                           & \cellcolor[HTML]{ED9445} 0.603           \\
XTREME                & 40   & 14   & \cellcolor[HTML]{ED9445}0.612   & \cellcolor[HTML]{EC8532}0.775 & \cellcolor[HTML]{F5DBA4} 0.311                           & \cellcolor[HTML]{F6E9B7} 0.457           \\
XGLUE                                                 & 19   & 7*   & \cellcolor[HTML]{F6E9B7}0.517   & \cellcolor[HTML]{F6E9B7}0.674 & \cellcolor[HTML]{F6E9B7} 0.307                           & \cellcolor[HTML]{F5DBA4} 0.504           \\
XNLI  & 15   & 7*   & \cellcolor[HTML]{F3CD91}0.557   & \cellcolor[HTML]{F2BE7E}0.711 & \cellcolor[HTML]{F0B06B} 0.339                           & \cellcolor[HTML]{EFA258} 0.598           \\
XCOPA                                                                     & 11   & 11   & \cellcolor[HTML]{EFA258}0.586   & \cellcolor[HTML]{EFA258}0.737 & \cellcolor[HTML]{EA771F} 0.361                           & \cellcolor[HTML]{EC8532} 0.608           \\
TyDiQA                                                 & 11   & 10   & \cellcolor[HTML]{EC8532}0.626   & \cellcolor[HTML]{ED9445}0.751 & \cellcolor[HTML]{ED9445} 0.343                           & \cellcolor[HTML]{F3CD91} 0.525           \\
XQuAD                                         & 12*  & 6*   & \cellcolor[HTML]{F5DBA4}0.523   & \cellcolor[HTML]{F5DBA4}0.680 & \cellcolor[HTML]{EFA258} 0.341                           & \cellcolor[HTML]{F0B06B} 0.588           \\
\hline
TeDDi                                                                                        & 89   & 51   & \textbf{0.706} & -                             & \textbf{0.369} & -              
\end{tabular}
\caption{Diversity of multilingual NLP data sets. The numbers in the second and the third column marked with an asterisk are added or modified by us. The numbers without an asterisk are reported in the respective publications. N(L): the number of languages in the data set. N(F): the number of families to which the languages belong. TI: typology index \citet{ponti-etal-2020-xcopa}. $J_{mm}$: Jaccard minmax similarity (this paper).
}
\label{tab:benchmarks}
\end{table*}

With the grammar features extracted from URIEL, we calculate syntactic diversity according to both TI and J$_{mm}$.

\smallskip
\noindent
\textbf{Syntax Typological Index (TI$_{syn}$)}
Following the formulation by \citet{ponti-etal-2020-xcopa}, we calculate the typological index for each data set. In this context, a language is characterized by 103 syntactic features with binary values\footnote{We use the syntax\_knn features available in lang2vec, which includes predicted values for those languages whose features are not available}. For each feature, Shannon entropy is estimated using the distribution of feature values in a data set. The feature-specific entropy values are averaged over the full set of features to obtain a TI score ranging from 0 to 1. The TI values closer to 1 indicate more diversity.


\smallskip
\noindent
\textbf{Syntax Jaccard ($J\_{mm\_syn}$)}
We apply Jaccard similarity for comparing each data set against the
TeDDi sample. Here the measures are the counts of the observed values of the same 103 syntactic feature available in \texttt{lang2vec}. This means that the items on the x-axis in Figure \ref{fig:ex_Jaccard} are the 103 values, while the y-axis represents the number of times each feature value was observed in a data set. Since these feature values are binary, the width of the bin is not arbitrary in this case; it is determined by the values. Conceptually, grouping several features into a single one would correspond to increasing the bin width, but it is not clear at the moment how the features could be grouped. We thus work with the original set without any changes. 

\bigskip

With text features (mean word length) extracted from TeDDI and the scored NLP data sets, we calculate morphological diversity according to both TI and J$_{mm}$. 

\smallskip
\noindent
\textbf{Morphology Typological index (TI$_{morph}$)} 
We adapt the measure proposed by \citet{ponti-etal-2020-xcopa} to the text-based features (mean word length). Each bin of the mean word length values is a feature and the number of languages that fall in a given bin are the counts of feature values. In other words, the mean word length becomes a vector of binary values, 1 for the languages that are in the bin and 0 for all the other languages in the sample. The rest of the calculation is the same as in TI$_{syn}$.


\smallskip
\noindent
\textbf{Morphology Jaccard ($J\_{mm\_morph}$)}
Similarly to $J\_{mm\_syn}$, we calculate the Jaccard score by comparing the distributions of the mean word length: TeDDi vs. a given NLP data set.




\section{Findings}
\label{sec:results}

\begin{figure*}
    \centering
    \includegraphics[width=1\textwidth]{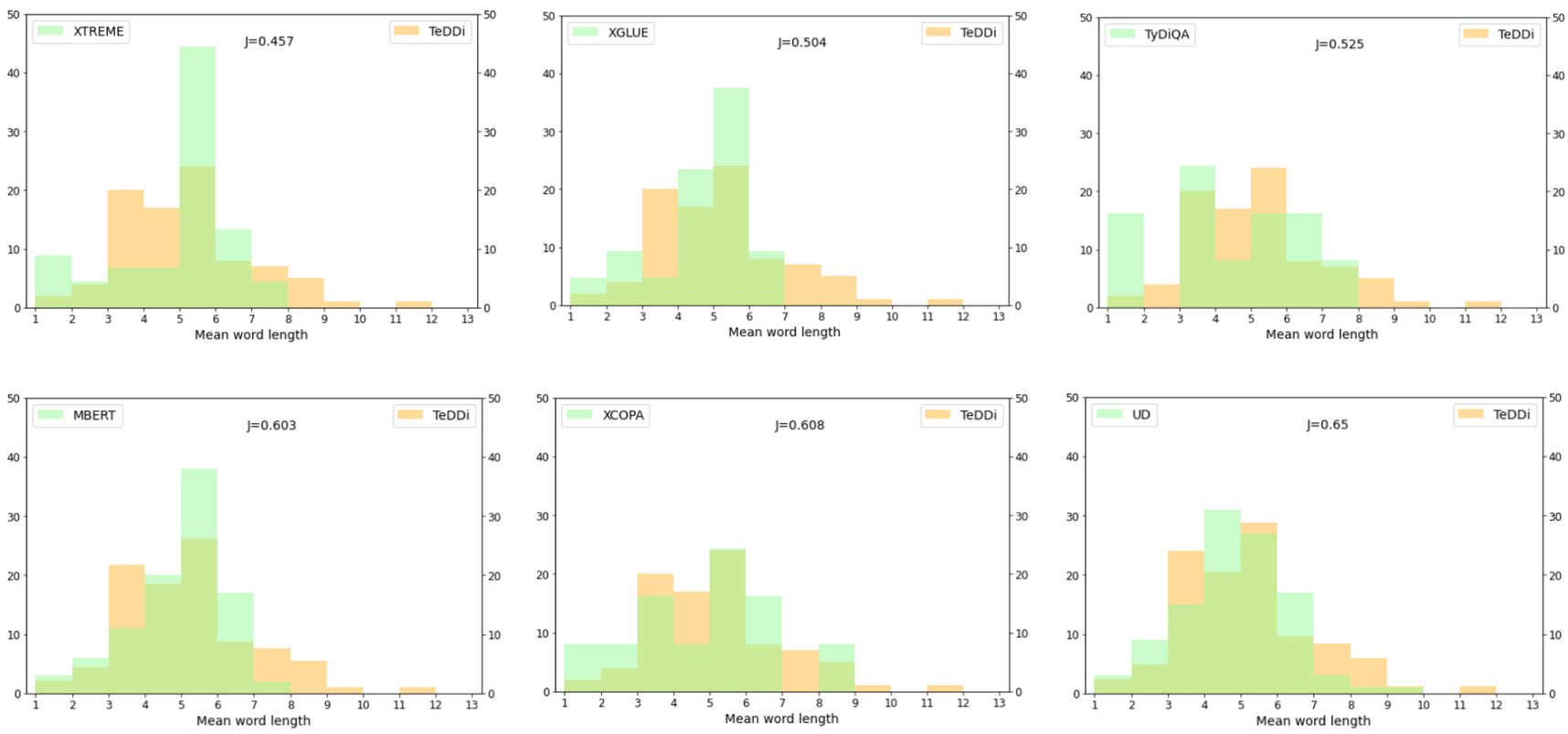}
    \caption{Union and intersection between the distributions of the mean word length in TeDDi and NLP data sets.}
    \label{fig:overlap_plots}
\end{figure*}

Table {\ref{tab:benchmarks}} lists all the reviewed data sets with all the measures of linguistic diversity. The colour scale of the cells represents the relative ranking of data sets according to each measure separately. 
TeDDI data set obtains the highest diversity scores at both levels (syntax and morphology) using the TI measure. This confirms the role of these resources as the current reference regarding linguistic diversity. 

\smallskip
\noindent
\textbf{TI and J$_{mm}$ are consistent} The rankings of data sets according to the J$_{mm}$ score are very similar to those obtained with the TI score when the syntactic features are used. The agreement between the two measures is somewhat lower in the case of morphological features, but still rather high. The consistency between the two measures is not a trivial outcome given the entirely different approaches behind them. We can thus take this agreement as a validation of both measures. The main advantage of J$_{mm}$ compared to TI is its transparency regarding the kinds of languages that are missing. The difference with respect to the reference is visible at the level of features indicating the values that need to be added or removed to improve the diversity. 

\smallskip
\noindent
\textbf{Diversity rankings of NLP data sets}  
The highest rankings appear split between the two structural levels. Bible 100 \citep{BIBLE100} and XTREME \citep{Hu2020xtreme} are the two most syntactically diverse data sets, while their morphological diversity is moderate to low. The Bible data set contains mostly non Indo-European languages, while the collection criteria for the XTREME data set was to maximise diversity.  
On the other hand, Universal Dependencies (UD, \citet{nivre-etal-2020-universal}, which are often seen as especially biased towards European languages, show the best morphological, but a moderate syntactic diversity. XCOPA \citep{ponti-etal-2020-xcopa} and TyDiQA \citep{clark-etal-2020-tydi} are  data sets containing relatively few languages, but designed to maximise linguistic diversity. They are both highly ranked on 3/4 measures (two syntactic and one morphological). Contrary to this, the linguistic diversity ranking of one of the most popular benchmarks that contain manual labels for several downstream tasks,  
XGLUE \citep{liang-etal-2020-xglue,wang-2019-glue-iclr} is consistently low. XQuAD \citep{artetxe-etal-2020-cross,rajpurkar-etal-2016-squad} fairs a little better, but it is still one of the least diverse data sets. The XNLI data set \citep{conneau-etal-2018-xnli,bowman-etal-2015-large,williams-etal-2018-broad}, which is compiled with the goal of spanning language families and which includes some low resource languages, remains of moderate linguistic diversity according to all measures.  It is curious to see that the number of languages or even languages families included in a data set does not ensure a high linguistic diversity. For example, the mBERT\footnote{\url{https://github.com/google-research/bert/blob/master/multilingual.md}} data set contains 97 languages in 15 language families, but it turns out to be less diverse than smaller data sets such as XCOPA (on TI$_{syn}$, J$_{mm\_syn}$ and J$_{mm\_morph}$) and TyDiQA (on TI$_{syn}$, J$_{mm\_syn}$ and TI$_{morph}$). The strategy of including the top 100 languages according to the size of their Wikipedia content (plus Thai and Mongolian), does not result in high diversity.

\smallskip
\noindent
\textbf{Underrepresented language types}
Figure \ref{fig:overlap_plots} is a visualisation of the J$_{mm\_morph}$ score\footnote{We show the morphological diversity for convenience since visualising 103 syntactic features would required additional adaptations.} for some of the data sets showing the overlap and differences with the reference (TeDDi). The recurrent difference is whether a data set includes languages with long words or not (mean length $> 8$). Those that contain at least some languages with long words (UD, XCOPA) score much better on $J_{mm\_morph}$ than those that remain completely on the short-middle side (EXTREME, XGLUE, TyDiQA, mBERT). 
The second important factor that leads to lower scores is a strong peak of the distribution indicating a bias towards one of the length bins (EXTREME, XGLUE, mBERT). The third factor is a different (``wrong'') shape of the distribution (TyDiQA). The data set that diverges the most is EXTREME, exhibiting all three factors of disagreement.  

The information about what kinds of languages are missing in a data set can be used to adjust language sampling and improve diversity. This is relatively straightforward when we deal with a single feature such as the mean word length. For example, the diversity of the mBERT language sample would be improved if the number of languages with a mean word length between 3 and 4 is reduced (by removing a given number of randomly selected languages). Instead of these languages, one should add a given number of languages with a mean word length greater than 7. It is not obvious where to look and how to find such languages (beyond the TeDDI sample), but knowing that they are needed might motivate such searches. Multi-feature scores (such as feature entropy) could specify the needed languages more precisely, but they would require an optimisation method to ensure that a newly added language increases indeed the diversity score. It might happen, for instance, that we want to increase the count on one feature value but not on another. In this case, we need a language that has 1 on the desired feature value but 0 on the features that we do not want to change. Devising such a method is beyond the scope of the current paper, but it is a clear next step for future work.

Overall, it seems that the right-hand side of the mean word length scale remains rather scarcely represented in all data sets, including the TeDDi sample itself. In future data collection, more effort should be put into representing languages with long words, especially because most of them are endangered. There are 12 languages in the TeDDI sample with a mean word length of over 7. If we localise them in Figure 1, we can see that ten of them are classified as extinct, endangered or vulnerable: Apurinã (apu), Chukchi (ckt), Kalaallisut (kal), Kayardild (gyd), Makah (myh), Martuthunira (vma), Plains Cree (crk), Ngiyambaa (wyb), Wichita (wic) and Yagua (yad). Only two of these languages, Luvale (lue) and Zulu (zul) are safe.

\section{Discussion}
\label{sec:discussion}

Our linguistic diversity scores include two kinds of language features (expert features extracted from data bases and the mean word length as a text feature) describing two structural levels (syntax and morphology). Readers not familiar with the details of how expert features are used in NLP might be left wondering whether the use of the mean word length is necessary and whether this measure is a good approximation of morphological types. 

Describing the use of expert features in NLP in Section  \ref{sec:features}, we note that the library \texttt{lang2vec} does not contain any morphological features, although these features are present in linguistic data bases. It is not clear why this is the case, but this means that morphological features are currently not used in NLP to assess linguistic diversity and the distances between languages. One possible reason for omitting morphological features could be the problem of sparsity, which would become even worse with these features leading to even more imputed values. For instance, if we want to study the distribution of 27 morphological features, only 34 languages will have a value for all these features. The values for the thousands of other languages would need to be imputed. This is the main reason why we propose to complement the existing sources of expert features with the mean word length as a value that can be easily calculated for any language on a small sample of text (500 tokens). 

To justify this proposal, we show that an independent measure of morphological complexity ($C_{WALS}$) and the mean word length are strongly correlated, but the score of 0.69 means that the agreement is not perfect.  A closer look into these two variables (Table \ref{tab:mwlccorr} in Appendix \ref{sec:appendixmorph}) points to the limitations of both measures, especially concerning the high values. For example, Turkish is the most complex language according to $C_{WALS}$, but its mean word length is well under 7. Although the correlation score is high and not due to chance, such aggregate measures remain approximations of the structural properties of languages. Nevertheless, these approximations are useful for tracking and improving linguistic diversity in data sets at the level of precision that is currently possible. Better approximations are certainly achievable in future work. Since our methods are general and can be applied to any set of features, any future improvements in representing linguistic structures can be easily integrated. 

\section{Conclusion} 
We have shown that the linguistic diversity of NLP data sets can be consistently assessed by two independent measures, TI (proposed in previous work) and J$_{mm}$ (proposed in this paper). Both of these measures show that a high number of languages and language families included in a data set is not sufficient to ensure linguistic diversity. 

To make the assessment of linguistic diversity automatic and rather simple, we show that text-based features such as the mean word length can be used as linguistic descriptors. These features can be easily calculated on very small text samples (of length of 500 tokens), overcoming the obstacles posed by the need to extract linguistic features from typological databases. 

An advantage of the J$_{mm}$ score over TI and other previous indicators of linguistic diversity is its capacity to show what kinds of languages are missing in a given data set in comparison to a reference. Assessing popular NLP data sets with this measure revealed that the most underrepresented languages are those with rich morphology. This kind of direct and transparent comparison can improve multilingual NLP coverage in the long run. 

\section*{Acknowledgements}

This research is partially supported by the Swiss National Science Foundation (SNSF) grants 176305 and PCEFP1\_186841. We thank the anonymous reviewers for their suggestions, which have improved the clarity of the paper.

\clearpage
\newpage

\section*{Limitations}

A limitation of our study is that the two levels of linguistic structures are represented with different features: syntax with expert features from linguistic data bases and morphology with mean word length as a text feature. Our results suggest that the two measures agree more at the level of syntax than at the level of morphology. To draw sound conclusions about the impact of the structural level on the agreement between the two measures, we would need both kinds of features for both levels. While we indirectly compare text and expert features at the level of morphology (via the correlation test), we do not propose syntactic features that could be extracted from text. We focused here on the current gap in the available linguistic features (the lack of morphological features in \texttt{lang2vec}), but devising and validating text-based syntactic features would deserve more attention in future work.


\bibliography{custom}
\bibliographystyle{acl_natbib}

\newpage

\appendix

\onecolumn

\section{Mean Word Length Correlation between Different Sample Size}
\label{sec:appendixcorr}

\label{sec:word_length}
\begin{figure}[h!]
 \centering
    \includegraphics[width=1\textwidth]{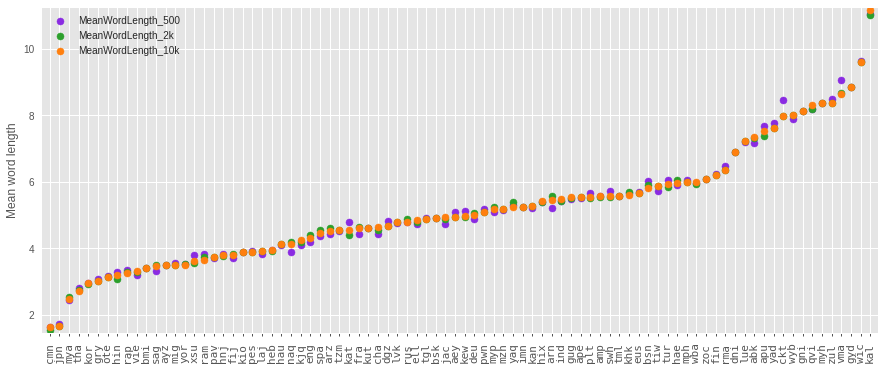}
    \caption{Mean word length measures at different text sizes in TeDDi. The languages on the x-axis are sorted according to the increasing value calculated on the biggest sample (10K). The values in the two smaller samples (2K and 500) depart very little from the main trend.}
    \label{fig:ex_mwl100LC}
\end{figure}

To make sure that the stability across different sample sizes suggested by Figure \ref{fig:ex_mwl100LC} is not a mere consequence of a relatively small range of variation, we perform correlation tests between different samples and in comparison to other measures (TTR and unigram entropy (H)). Table \ref{tab:sample_corr} shows that the ranks of languages change considerably less across different sample sizes when considering the mean word length than in the other two measures.  

\begin{table}[h!]
\centering
\begin{tabular}{|l||r|r|r|}
\hline
{Samples} & MWL & H & TTR \\
\hline
\hline
500 tokens vs. max. & 0.99 & 0.85 & 0.84 \\
\hline
2K tokens vs. max & 0.99 & 0.95 & 0.94\\

\hline
\end{tabular}
\caption{Spearman rank correlation showing how much rankings of languages change with text measures taken on random samples of different size.}
\label{tab:sample_corr}
\end{table}

\newpage

\section{Word length and morphological complexity}
\label{sec:appendixmorph}

\begin{table}[h!]%
\begin{tabular}{llll}
\hline
     ISO396-3   & Name &   MWL  &   $C_{WALS}$   \\

\hline
abk & Abkhazian &  7.17& 0.62\\
apu &  Apurinã & 7.67& 0.60\\
arz & Egyptian Arabic &  4.44& 0.49\\
bsn & Barasana-Eduria  &  6.02& 0.69\\
ckt &  Chukchi &  8.45& 0.50\\
deu &  German &  4.87& 0.55\\
ell & Modern Greek  &  4.72& 0.53\\
eng &  English &  4.18& 0.42\\
eus &  Basque &  5.70& 0.64\\
fin & Finnish  &  6.23& 0.66\\
fra & French  &  4.41& 0.45\\
hae &  Eastern Oromo &  5.91& 0.53\\
hau &  Hausa &  4.08& 0.38\\
heb &  Modern Hebrew &  3.94& 0.54\\
ind & Indonesian  &  5.42& 0.40\\
kan & Kannada  &  5.22& 0.65\\
kat &  Georgian &  4.78& 0.50\\
khk & Halh Mongolian  &  5.66& 0.53\\
kut &  Kutenai &  4.60& 0.37\\
lvk &  Lavukaleve &  4.77& 0.67\\
qvi &  Imbabura Highland Quichua &  8.18& 0.71\\
rus & Russian  &  4.79& 0.52\\
spa & Spanish  &  4.37& 0.45\\
swh &  Swahili &  5.72& 0.71\\
tur &  Turkish &  6.07& 0.76\\
vie &  Vietnamese &  3.20& 0.21\\
yaq &  Yaqui &  5.31& 0.57\\
yor & Yoruba  &  3.52& 0.25    \\
\hline
Spearmann correlation& $\rho = 0.69$

\end{tabular}

\caption{Mean Word length (MWL) and morphological complexity measure ($C_{WALS}$) in the subset of TeDDi languages for which 26 WALS morphology features are known.}
\label{tab:mwlccorr}
\end{table}

\begin{table}[h!]
\centering
\begin{tabular}{lp{5cm}lcc}
\hline

\textbf{Chapter}&	\textbf{Name}&	\textbf{Categories}&	\textbf{Transformation}&	\textbf{Final Values}\\
\hline
22A&	Inflectional Synthesis&	7 (ordinal)&	none&	1-7\\
26A&	Prefixing vs. Suffixing in Inflectional Morphology&	6 (non-ordinal)&	binarization&	0-1\\
27A&	Reduplication&	3 (non-ordinal)&	binarization&	0-1\\
28A&	Case Syncretism&	4 (ordinal)&	reorder&	1-4\\
29A&	Syncretism in Verbal Person/Number marking&	3 (ordinal)&	none&	1-3\\
30A&	Number of Genders&	5 (ordinal)&	none&	1-5\\
33A&	Coding of Nominal Plurality&	9 (partially ordinal)&	binarization&	0-1\\
34A&	Occurrence of Nominal Plurality&	6 (ordinal)&	none&	1-6\\
49A&	Number of Cases&	9 (ordinal)&	remove&	1-8\\
51A&	Position of Case Affixes&	9 (non-ordinal)&	binarization&	0-1\\
57A&	Position of Pronominal Possessive Affixes&	4 (non-ordinal)&	binarization&	0-1\\
59A&	Possessive Classification&	4 (ordinal)&	none&	1-4\\
65A&	Perfective/Imperfective Aspect&	binary&	none&	0-1\\
66A&	The Past Tense&	4 (ordinal)&	reorder&	1-4\\
67A&	The Future Tense&	binary&	none&	0-1\\
69A&	Position of Tense/Aspect Affixes&	5 (non-ordinal)&	binarization&	0-1\\
70A&	The Morphological Imperative&	5 (partially ordinal)&	recategorization&	1-4\\
73A&	The Optative&	binary&	none&	0-1\\
74A&	Situational Possibility&	3 (non-ordinal)&	binarization&	0-1\\
75A&	Epistemic Possibility&	3 (non-ordinal)&	binarization&	0-1\\
78A&	Coding of Evidentiality&	6 (non-ordinal)&	binarization&	0-1\\
94A&	Subordination&	5 (non-ordinal)&	binarization&	0-1\\
101A&	Expression of Pronominal Subjects&	6 (non-ordinal)&	binarization&	0-1\\
102A&	Verbal Person Marking&	5 (partially ordinal)&	recategorization&	1-3\\
111A&	Nonperiphrastic Causative Constructions&	4 (non-ordinal)&	binarization&	0-1\\
112A&	Negative Morphemes&	6 (non-ordinal)&	binarization&	0-1 \\
\hline
\end{tabular}
\caption{\label{tab:WALS} Subset of WALS features that we use for characterizing the morphological complexity of languages. The column ``Final Values'' gives the range of values each feature can take after transformations were performed to the original values \citep{bentz2016} }
\end{table}

\clearpage

\newpage

\section{Word Length Adjustments for Logographic Scripts}
\label{sec:logographic_adjustments}

\begin{table}[h!]
\centering
\begin{tabular}{p{6cm}rr|rrrr}
Name and main references                                                                                                                            & N(L) & N(F) & TI$_{syn}$                      & $J_{mm\_syn}$                 & TI$_{morph}$                    & $J_{mm\_morph}$ \\ \hline
Universal Dependencies (UD)                    & 106* & 20*  & 0.567   & 0.736 &  0.337                           &  0.665           \\
Bible 100                                                 & 103* & 30*  & 0.649   & 0.811 &  0.302                           &   0.617          \\
mBERT                                                & 97*  & 15*  & 0.559   & 0.710 &  0.316                           &  0.617           \\
XTREME                & 40   & 14   & 0.612   & 0.775 &  0.311      &  0.471           \\
XGLUE               & 19   & 7*   & 0.517   & 0.674 &  0.297     &  0.580           \\
XNLI  & 15   & 7*   & 0.557   & 0.711 &  0.321                           &  0.704     \\
XCOPA    & 11   & 11   & 0.586   & 0.737 & 0.336   &  0.634           \\
TyDiQA    & 11   & 10   & 0.626   & 0.751 & 0.343   & 0.552         \\
XQuAD    & 12*  & 6*   & 0.523   & 0.680 &  0.318   & 0.634         \\
\hline
TeDDi                                                                                        & 89   & 51   & \textbf{0.706} & -                             & \textbf{0.361} & -              
\end{tabular}
\caption{Diversity of multilingual NLP data sets with adjustments for logographic scripts. Compared to the main results in Table 1, all TI$_{morph}$ scores are slightly decreased and J$_{mm\_morph}$ slightly increased. The rankings of the t are mostly preserved, with the exception of XNLI, whose J$_{mm\_morph}$ ranking improves.  
}
\label{tab:benchmarks_logo}
\end{table}

\end{document}